\documentclass{article}
\usepackage{spconf}

\usepackage{graphicx}
\usepackage{amssymb}
\usepackage[utf8]{inputenc} 
\usepackage[T1]{fontenc}    
\usepackage{url}            
\usepackage{booktabs}       
\usepackage{amsfonts}       
\usepackage{nicefrac}       
\usepackage{microtype}      
\usepackage{xcolor}         
\usepackage{tikz}
\usetikzlibrary{shapes.geometric}
\usepackage{subcaption}
\usepackage{float}
\usepackage{bm}
\usepackage{amsmath}
\usepackage[normalem]{ulem}
\usepackage{lipsum}
\useunder{\uline}{\ul}{}
\usepackage[pagebackref,breaklinks,colorlinks]{hyperref}

\newtheorem{knowledge}{Knowledge}
\begin{document}
\title{Towards Commonsense Knowledge based Fuzzy Systems for \\ Supporting Size-Related Fine-Grained Object Detection\\}
\name{Pu Zhang$^{\dag}$, Tianhua Chen$^{\sharp}$, Bin Liu$^{\dag,\ast}$
\thanks{$^\ast$Address correspondence to bins@ieee.org. This work was supported by Exploratory Research Project (No.2022RC0AN02) of Zhejiang Lab.}
}
\address{$^{\dag}$ Zhejiang Lab\\$^{\sharp}$ School of Computing and Engineering, University of Huddersfield}

%

\maketitle

\begin{abstract}
Deep learning has become the dominating approach for object detection. To achieve accurate fine-grained detection, one needs to employ a large enough model and a vast amount of data annotations. In this paper, we propose a commonsense knowledge inference module (CKIM) which leverages commonsense knowledge to assist a lightweight deep neural network base coarse-grained object detector to achieve accurate fine-grained detection. Specifically, we focus on a scenario where a single image contains objects of similar categories but varying sizes, and we establish a size-related commonsense knowledge inference module (CKIM) that maps the coarse-grained labels produced by the DL detector to size-related fine-grained labels. Considering that rule-based systems are one of the popular methods of knowledge representation and reasoning, our experiments explored two types of rule-based CKIMs, implemented using crisp-rule and fuzzy-rule approaches, respectively. Experimental results demonstrate that compared with baseline methods, our approach achieves accurate fine-grained detection with a reduced amount of annotated data and smaller model size. Our code is available at: https://github.com/ZJLAB-AMMI/CKIM.
\end{abstract}

\begin{figure*}[ht]
\centering
\includegraphics[width=0.35\textwidth]{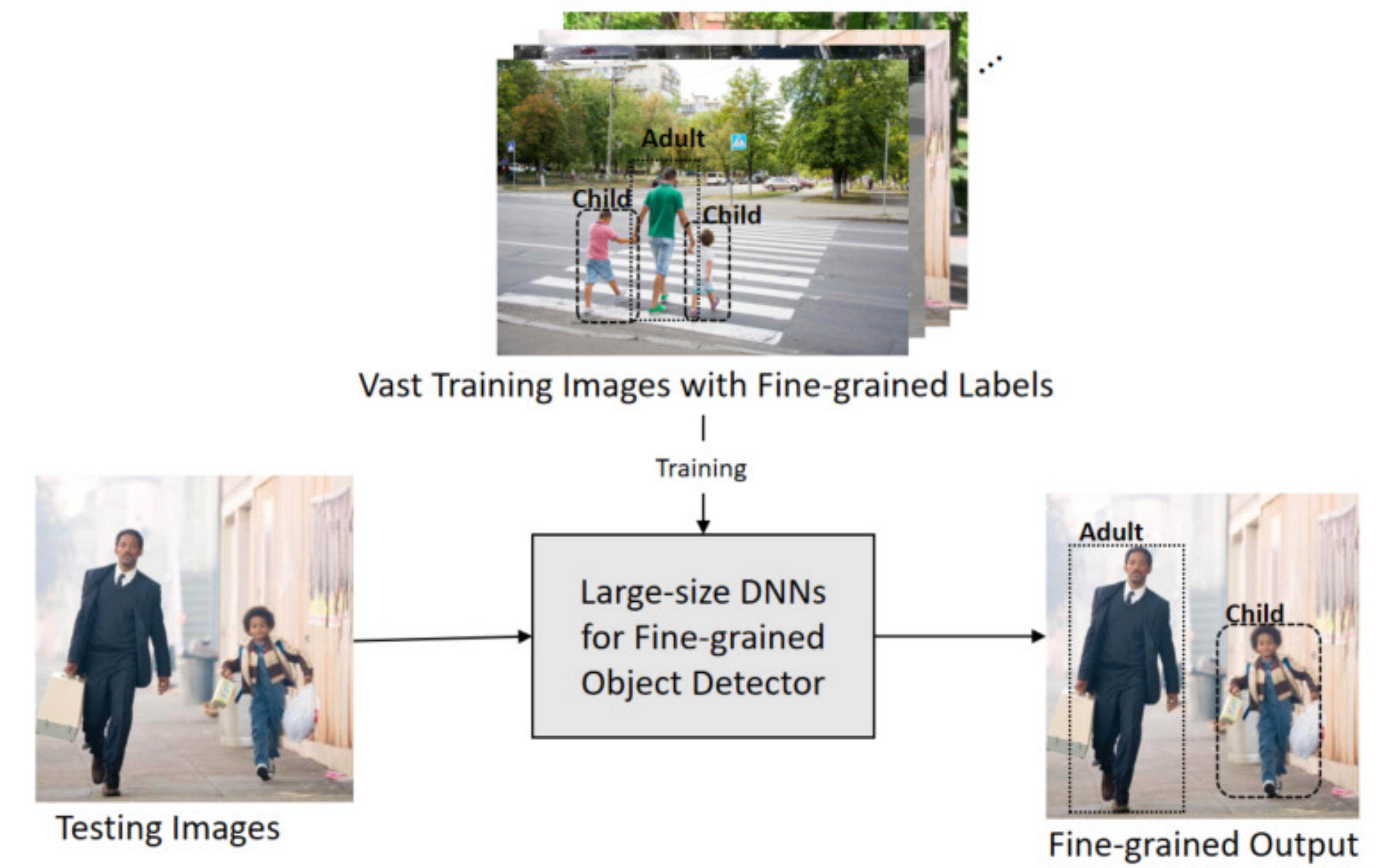} \quad \quad \quad \includegraphics[width=0.48\textwidth]{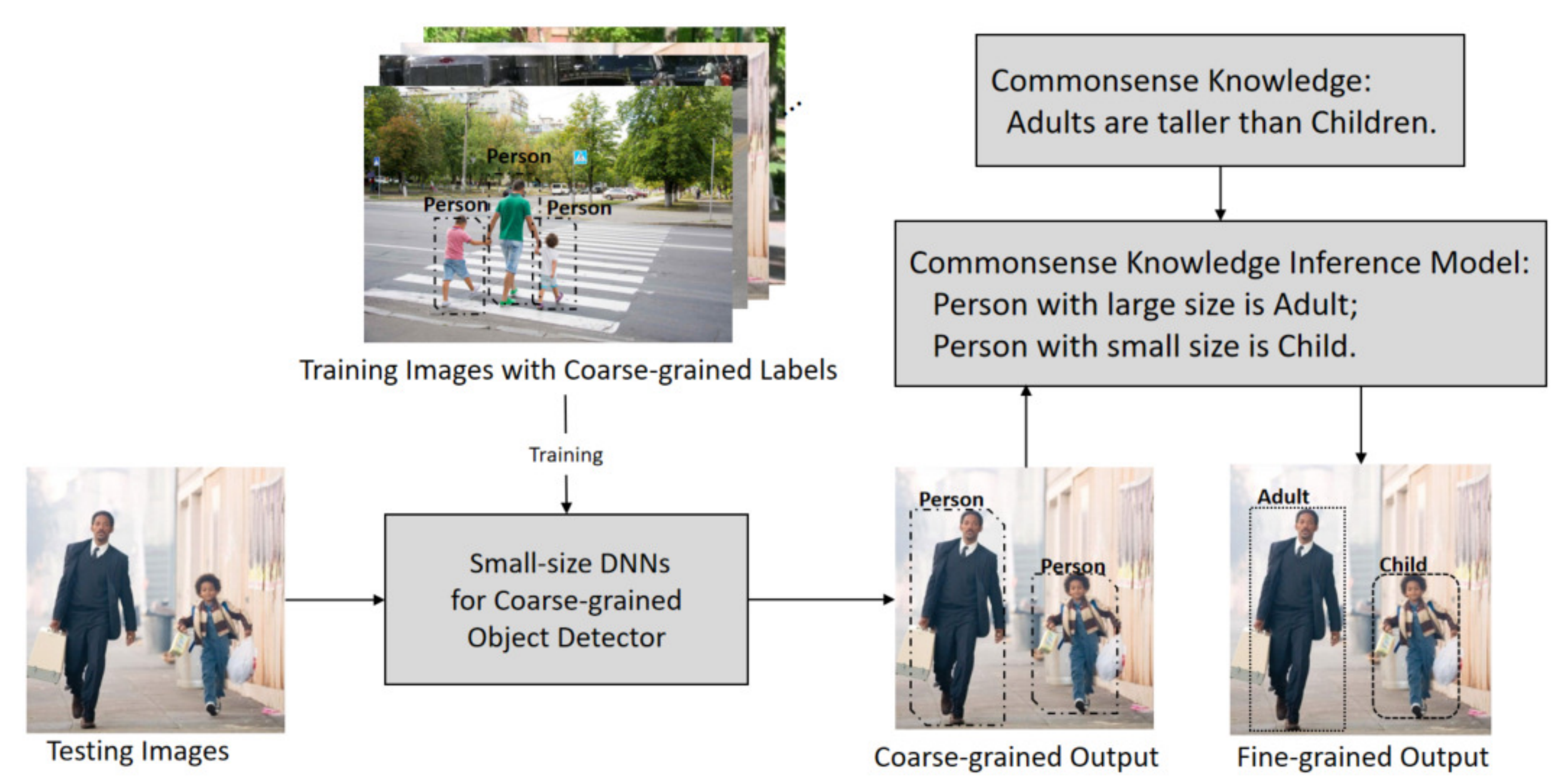}
\caption{Comparison of the working mechanisms between a typical fine-grained object detector (left) and our approach (right)}
\label{fig:general_model}
\end{figure*}

\section{Introduction}

Size-related fine-grained object detection appears in many real-life scenarios. For example, in autonomous driving, we want the algorithm to be capable of automatically identifying individuals as well as their size-related attributes. This is critical for the system to discriminate between adults and children, as the driving strategies for adults and children are likely to be different. When we request a robot to fetch a cup of an specific size for us, the robot not only needs to recognize the presence of cups in the scene, but also accurately measure the size of each cup to achieve the desired task.


Modern deep learning (DL) based methods, such as YOLO \cite{redmon2018yolov3} and Faster R-CNN \cite{ren2015faster}, have become the dominant approach to object detection as it significantly improves detection accuracy. However, the benefits of these advancements come at a cost: the requirement on substantial computational and storage resources, along with a vast quantity of annotated data. This challenge is further compounded in the fine-grained case of our concern, which demands even larger models, more annotated data, and greater computational budgets to train and store the model \cite{zaidi2022survey}.

In this paper, we propose a lightweight deep neural network (DNN)-based detector for fine-grained objects that are size-related. When presented with an image containing multiple objects of similar category but varying in size, our approach can efficiently determine the category and size of each object. The core idea is to leverage commonsense knowledge, independent of specific tasks, to enhance the performance of a coarse-grained object detector. To achieve this, we identify two valuable pieces of size-related commonsense knowledge and introduce a commonsense knowledge-based inference module (CKIM) that collaborates with the coarse-grained object detector. The CKIM can infer fine-grained object labels based on the output generated by the coarse-grained detector. Fig.\ref{fig:general_model} illustrates the distinctions between traditional fine-grained detectors and our proposed approach.


To harness the proposed commonsense knowledge, we plan to employ knowledge-based systems. These systems will explicitly encode this knowledge using tools like production or IF-THEN rules. IF-THEN rules are advantageous as they clearly define judgements based on specific conditions, enabling the system to logically deduce conclusions and explain its reasoning process to users \cite{chen2021decision}. However, traditional crisp rule-based knowledge representation often involves setting strict thresholds for variables. While effective in some scenarios, this rigid approach struggles with ambiguous and vague concepts essential in practical reasoning and decision-making. In our study, such concepts include the notions of distance and size, which are central to our commonsense knowledge framework for inference and reasoning.

Rather than relying on the rigid structure of crisp rules, fuzzy systems are gaining recognition for their effectiveness in developing knowledge-based systems, especially in scenarios where data or knowledge is inherently imprecise. Fuzzy set-based systems are particularly suitable for the underlying application as they accommodate the ambiguity and imprecision typical of many real-world concepts. These systems enable a form of reasoning that more closely resembles human thought processes, adept at handling expressions like "large object" or "short distance" that are often used to describe entities. This approach not only enhances the transparency of the models themselves but also improves the clarity of the inferences made by these models. Numerous methodologies have been explored for creating and training fuzzy knowledge-base systems, demonstrating their versatility and effectiveness across a wide range of practical applications \cite{carter2021fuzzy}.

Building on the encouraging findings in existing research, our goal is to leverage commonsense knowledge in enhancing size-related fine-grained object detection. This paper outlines a plan to develop knowledge-based systems that incorporate this common sense knowledge. While the advantages of using fuzzy systems are clear from our previous analysis, we intend to adopt a more empirical approach. Our strategy involves conducting a comparative study that develops both a crisp rule-based system and a fuzzy rule-based system. This comparison will help us determine the most effective knowledge-based system to augment a lightweight deep neural network-based coarse-grained object detector.

\section{Related Works}\label{sec:related}

\subsection{Modern DNNs for Visual Object Detection}
There are two major types of modern DNN based object detectors. One is termed two-stage detectors, such as Faster R-CNN \cite{ren2015faster}, which include networks with separate modules for candidate region generation and classification. The second type is single-stage detectors, which directly produce object categories and their bounding boxes in a single step and use pre-defined differently sized boxes to locate objects. Single-stage detectors are suitable for real-time, resource-constrained scenarios such as edge computing because they have lightweight designs and require less time to make predictions. So we employ them as benchmark object detectors in this paper.

You Only Look Once (YOLO) is one of the most widely used single-stage object detectors \cite{redmon2016you}. In YOLO, the input image is divided into a grid of cells, and each cell predicts bounding boxes and their corresponding class probabilities. Over time, newer versions of YOLO, such as YOLOv7, have been proposed to improve inference speed and reduce model size. YOLOv7 have lightweight versions, namely YOLOv7-tiny, which employ smaller model architectures that are better suited for edge computing scenarios \cite{wang2021scaled,wang2022yolov7}. In our experiments, we use YOLOv7-tiny as a baseline method.

MobileNet-SSD is an efficient DL method for object detection in mobile computing scenarios. It employs a Single Shot MultiBox Detector (SSD) \cite{liu2016ssd} as the decoder and MobileNet \cite{howard2017mobilenets} as the feature extractor. SSD was the first single-stage detector that performed comparably to two-stage detectors. MobileNet was specifically designed for vision applications on mobile devices, with traditional convolution layers replaced by depthwise separable and pointwise convolution layers to reduce model size \cite{sandler2018mobilenetv2}. An advanced version of MobileNet, MobileNetV3, uses a search algorithm to optimize the network architecture \cite{howard2019searching}. In our experiments, we use MobileNetV3-SSD as a baseline method.

\subsection{Rule Based Systems}


Rule-based systems provide the computational mechanisms that can be found in most expert or knowledge-based systems \cite{KURFESS2003609}. The knowledge bases are represented as a collection of rules that are typically expressed as if-then clauses. In constructing such rule-based systems, the conventional approach has been to use crisp rule-based knowledge representation and strict thresholds. Typically, during the inference process, each individual instance can only fired at most one rule. When dealing with uncertain information or imprecision linguistic term, conventional crisp rule-based systems may not be effective. Fortunately, with the support of fuzzy logic and fuzzy set theory, such ambiguity and vague variables can be precisely describe by fuzzy sets.

The inference mechanism of most fuzzy rule based systems(FRBSs) follows the compositional rule of inference (CRI) principle \cite{zadeh1973outline}. That is, if the input coincides with the antecedent of a fuzzy rule, then the output should coincide with the corresponding consequent of that fuzzy rule. As such, FRBSs allow an instance to simultaneously match with multiple rules, as long as the instance overlaps with corresponding rule antecedents. The inference is achieved by integrating the conclusions derived from all overlapped rules. Mamdani models \cite{Scherer2012} are one of most widely used model due to their effectiveness in addressing classification and prediction problems in continuously valued domains. The antecedents and consequents of Mamdani rules are both represented by fuzzy sets. A defuzzification process is usually required to obtain crisp results in practice. 


\subsection{Neural-Symbolic Approaches}
Neural-symbolic approaches aim to integrate the complementary strengths of machine learning and symbolic reasoning \cite{yang2022neurosymbolic,garcez2022neural,yu2023survey}. There are different ways to combine DL and symbolic knowledge. One approach involves translating symbolic knowledge into continuous vector representations and treating them as inputs for DL models \cite{mikolov2013distributed,pennington2014glove}. Another approach is to instruct the training process of DL models with symbolic knowledge, such as by embedding the knowledge into the loss function \cite{nickel2016holographic}. Finally, it is also possible to perform symbolic reasoning based on the outputs of DL algorithms \cite{yi2018neural}\cite{eiter2022neuro}.

Our approach falls into the category of performing symbolic reasoning based on the outputs of DL algorithms. However, it differs from other methods in this category in several ways. Firstly, our target task is different. Secondly, the knowledge we employ is distinct from what has been used in prior neural-symbolic approaches. In particular, we identify and utilize two specific pieces of commonsense knowledge that are unique to our approach. Lastly, we develop novel techniques to effectively combine our knowledge with the outputs of modern coarse-grained object detectors.

\section{Size-related Commonsense Knowledge}\label{sec:CKIM}

\begin{figure}[ht]
\centering
\includegraphics[width=0.4\textwidth]{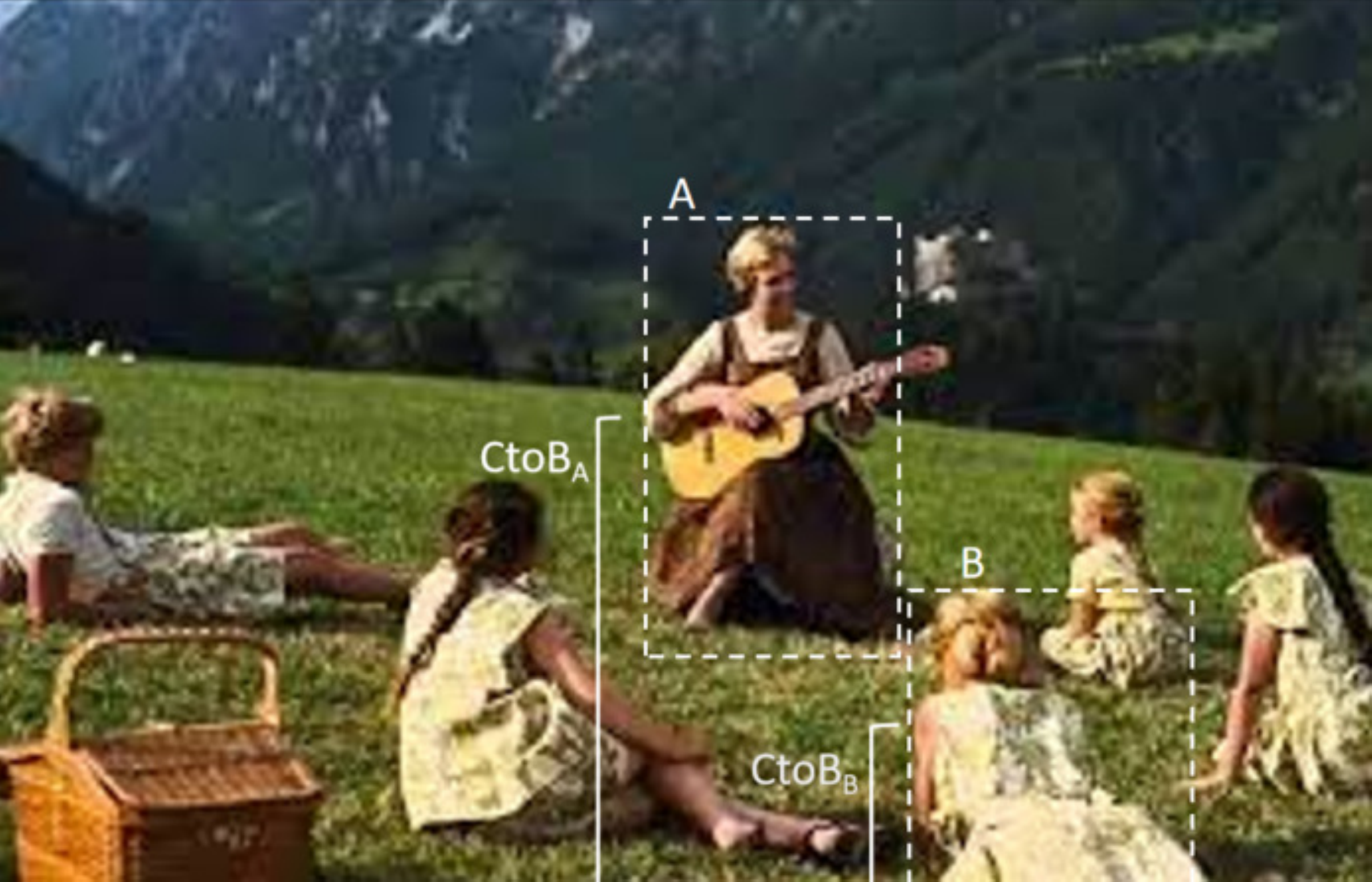}
\caption{An example image that illustrates the connection between an object's distance to the camera (DtoC) and the distance from the center of the object to the bottom of the image (CtoB). As shown in the image, CtoB of object A, denoted by CtoB$_A$, is larger than CtoB$_B$. This information can be used to infer that DtoC of object A, denoted by DtoC$_A$, is larger than DtoC$_B$, which is clearly true as shown in the picture.}
\label{fig:distance}
\end{figure}

In this section, we present our approach for size-related fine-grained object detection. As shown in the right panel of Fig.\ref{fig:general_model}, our approach consists of two parts: a coarse-grained object detector and CKIM. The former part outputs coarse-grained labels and a bounding box for each object. The bounding box can be represented by:

\begin{equation}
box(C,X,Y,W,H)
\end{equation}
where $C$ denotes the coarse-grained label, $(X,Y)$ denotes the coordinates of the centre of the bounding box, and $W$ and $H$ denote the width and height of the bounding box, respectively. We focus on size-related fine-grained labels in this work. The real size of an object is naturally closely related to the size of the object's bounding box, which can be produced by a qualified object detector. In addition, according to commonsense knowledge as follows:
\begin{knowledge}
An object appears to decrease in size as it moves farther away and appears to enlarge as it moves closer to the observer,
\end{knowledge}
\noindent
the real size of an object is related to its distance from the camera. For an object in an image, we can estimate its real distance to the camera based on its distance from the center of its bounding box to the bottom of the image. This is because:
\begin{knowledge}
the object's distance to the camera (DtoC) is strongly connected to the distance from the center of its bounding box to the bottom of the image (CtoB).
\end{knowledge}
\noindent The connection between CtoB to DtoC is illustrated in Fig.\ref{fig:distance}.

These two commonsense knowledge provides us with an opportunity to reason size-related fine-grained labels from the outputs of a coarse-grained object detector, such as the coarse-grained labels and bounding boxes of the objects.
The purpose of CKIM is to infer fine-grained labels based on the size of the objects' bounding boxes (BoxS) and their distances to the camera (DtoC). To achieve this, we normalize values of $X$, $Y$, $W$, $H$ according to the width and height of the image. Then, we calculate BoxS and DtoC as follows:  
\begin{equation}
\begin{aligned}
BoxS &= W \times H \\
DtoC &= 1-Y
\end{aligned}
\end{equation}


\begin{figure*}[h]
\centering
\includegraphics[width=0.8\textwidth]{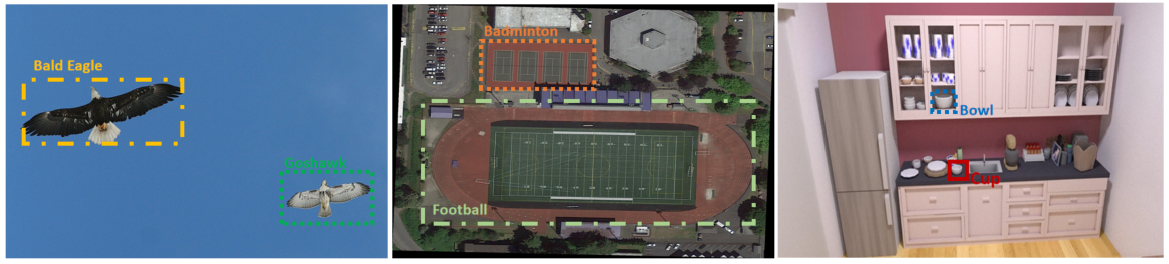}
\caption{Examples images for which the camera position for photographing the objects is located directly above or directly below the objects. Left: image of flying birds; Middle: remote sensing images; Right: Bowl in the cupboard and cup on the counter.}
\label{fig:case}
\end{figure*}

\subsection{On the Validity of Our Selected Commonsense Knowledge}\label{sec:validity}
To test the validity of the two commonsense-knowledge, we conducted an brief investigation on 10 popular benchmark datasets in 4 scenarios: autonomous driving, indoor robots, life scenes, and remote sensing. Specifically, we randomly selected 500 images which contain size-related categories, such as children and adults, car and bus, football and tennis fields, and check whether the two pieces of commonsense-knowledge are valid. The results of our investigation is summarized in Table \ref{tab:efficiency}. It turns out that Knowledge 1 consistently holds true. Moreover, Knowledge 2 is confirmed to be valid on all three autonomous driving datasets considered, with a probability exceeding 98\% for life scene image datasets, and with a probability surpassing 95\% for two indoor robot-related datasets. However, Knowledge 2 does not hold for two remote sensing image datasets.

As shown in Fig.\ref{fig:case}, we revealed that when the objects are positioned within the same plane perpendicular to the capture direction, the true size of the object can be directly determined by the size of their bounding boxes, without the need for knowledge 2 as shown in Fig.\ref{fig:case}.
\begin{table}[h]
\small
\centering
\begin{tabular}{llcc}
\toprule
\multicolumn{2}{c}{Datasets}   &  Knowledge 1  & Knowledge 2 \\
\hline
Autonomous  & BDD 100K & 100\% & 100\% \\
\cline{2-4}
Driving& CityPersons & 100\%  & 100\% \\
\cline{2-4}
&KITTI & 100\% & 100\%  \\
\hline
Indoor  & Habitat AI & 100\% & 95\% \\
\cline{2-4}
Robots & OpenLORIS & 100\% & 96\% \\
\hline
Remote  & WHU-RS19 & 100\% & 0\% \\
\cline{2-4}
Sensing &  RSSCN7  & 100\% & 0\%  \\
\hline
Life Scenes  & MSCOCO & 100\% & 99\% \\
\cline{2-4}
    & LVIS & 100\% & 98\%  \\
\cline{2-4}
&PASCAL VOC& 100\% & 99\%\\
\bottomrule
\end{tabular}
\caption{The validity of these two commonsense knowledge on 10 benchmark datasets}
\label{tab:efficiency}
\end{table}




\begin{figure}[h]
\centering
\includegraphics[width=0.4\textwidth]{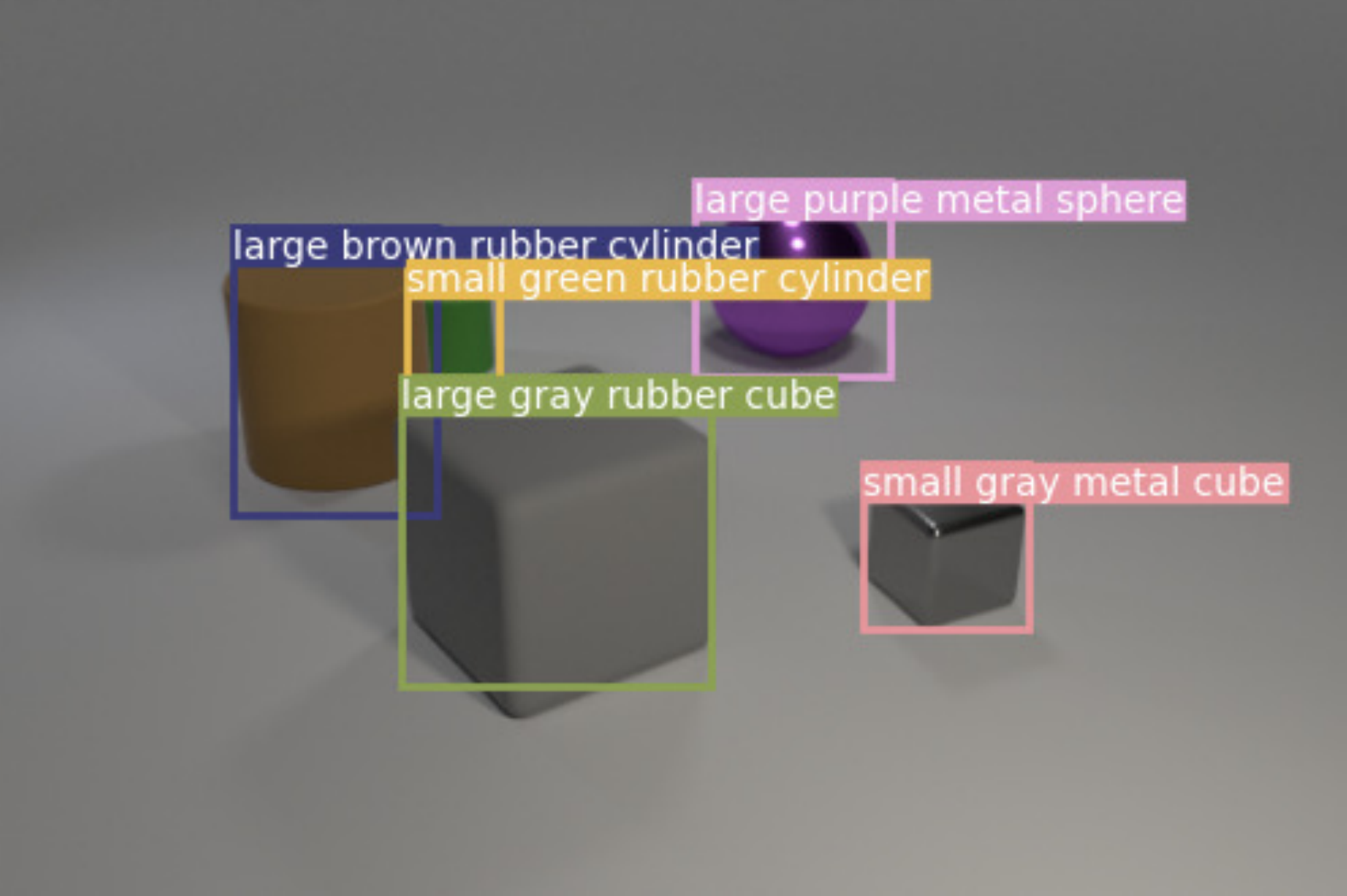} \\ \includegraphics[width=0.4\textwidth]{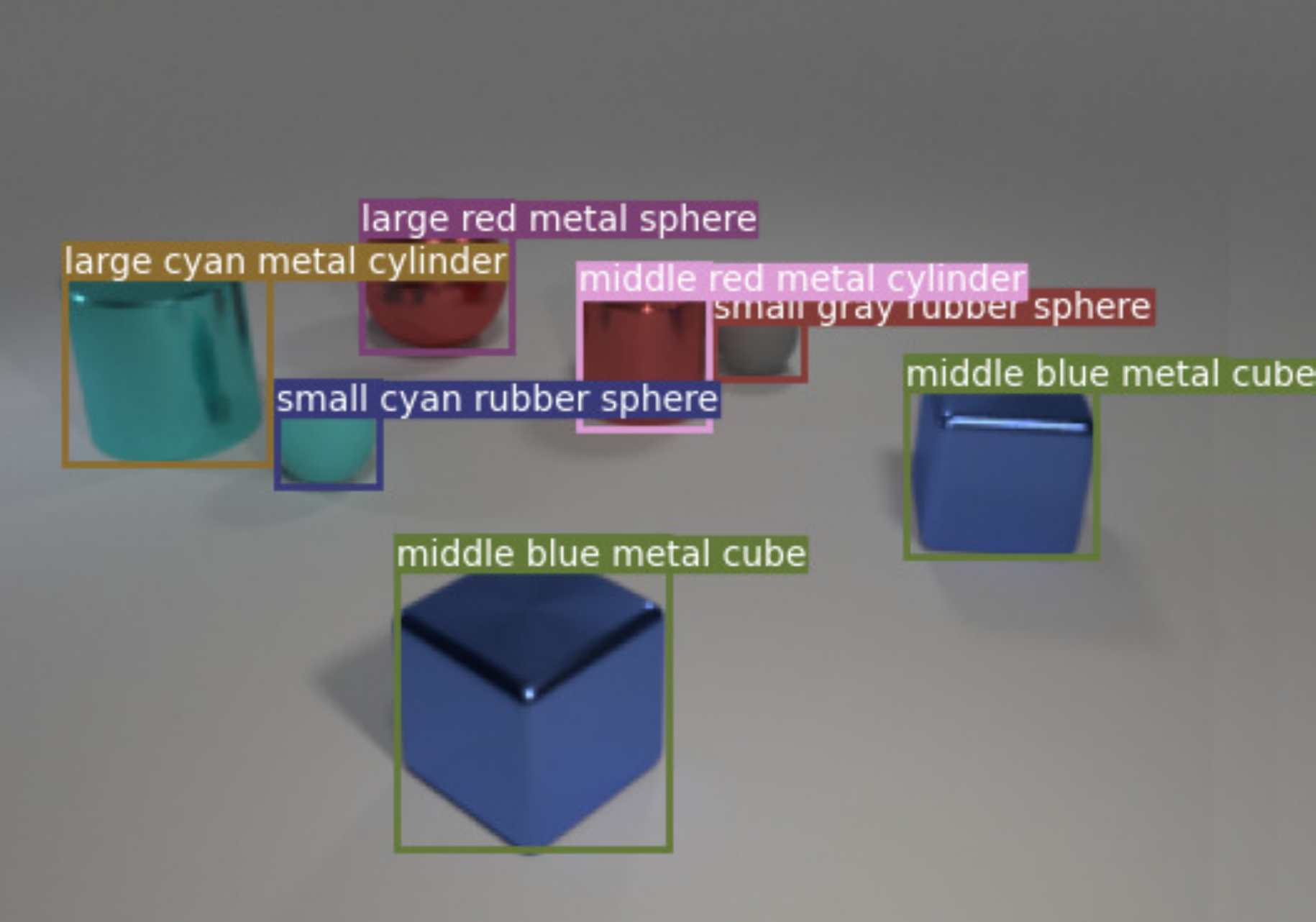}
\caption{Top: an example image in the CLEVR-96 dataset, where the object size attribute is specified as either `large' or `small'; Bottom: an example image in the CLEVR-144 dataset, where the object size attribute can be `large', `middle', or `small'.}
\label{fig:CLEVR}
\end{figure}

\section{CKIM Assisted Object Detection}\label{sec:CKIM}

In this section, we explain how to specify CKIM using commonsense knowledge and how to use it to assist a coarse-grained deep detector in doing fine-grained object detection. We develop two types of CKIM, corresponding to crisp-rule and fuzzy-rule based inference.

\subsubsection{Crisp-rule based CKIM}
In crisp rule-based systems, Boolean logic is followed, and an object can only belong to one class or not belong to it. We model the relationship between the real size of an object and its attributes given by a coarse-grained object detector using a logistic regression function \cite{lavalley2008logistic}.

To simplify the presentation, we consider size-related object labels: `large', `middle', and `small'. We use the one-vs-rest method to perform crisp rule inference. The decision function of the crisp rule is defined as follows:

\begin{equation}
f_{i}(x) = \frac{1}{1+e^{w_{i}^{T}x}}, i\in \{ml,sm\}
\end{equation}
where $x \triangleq (BoxS, DtoC)$, $f_{ml}$ the decision function for middle sized and large sized objects, and  $f_{sm}$ denotes the decision function for small sized and middle sized objects, $w_{i}^T$ is a set of parameters optimized by minimizing a loss function with stochastic gradient descent (SGD) algorithm. The loss function is defined as follows:

\begin{equation}
J(w_{i})=\sum_{(x,y)\in D}[-y\log(f_{i}(x))-(1-y)\log(1-f_{i}(x)]
\end{equation}
where $(x,y)$ represents input variables and the corresponding labels in dataset $D$. Accordingly, the CKIM is defined as follows:
\\

If $f_{ml}(x) > 0.5$,

\hspace{2em} then the object's fine-grained label is `large';

If $f_{ml}(x) < 0.5$ and $f_{sm}(x) > 0.5$,

\hspace{2em} then the fine-grained label is `middle';

If $f_{sm}(x) < 0.5$,

\hspace{2em} then the fine-grained label is `small'.
\\
\subsubsection{Fuzzy-rule based CKIM}
In real-life cases, there may be continuous-valued attributes (e.g., size, distance, age) that are difficult to classify with discrete semantic descriptions (e.g., large or small, near or far, young or old), resulting in semantic vagueness or uncertainty. Fuzzy logic provides a multi-value logic, in which such vague semantics can be strictly formulated and precisely studied.

To adapt our method to more general scenarios, we propose a fuzzy-rule based approach for implementing CKIM. Fuzzy rule-based systems allow an object to match different categories with different memberships. We adopt the aforementioned Mamdani model for this task, which uses fuzzy sets as rule antecedents and consequents. Two main categories of membership functions are typically used in constructing fuzzy sets \cite{2021Approximate}:

(1) Polygonal functions, such as triangular shaped and trapezoidal shaped functions;

(2) Nonlinear functions, including Gaussian shaped and generalised bell shaped functions.

Here, we adopt the Gaussian-shaped membership function \cite{kreinovich1992gaussian}. As before, we consider size-related object labels, namely `large', `middle', and `small'.

In our fuzzy rule inference module, the antecedents of rules are defined as follows:
\begin{equation}
\begin{aligned}
M_L(x) &= \mathcal{N}(x|\mu_L,\Sigma_L)\\
&=\frac{1}{2\pi|\Sigma_L|^{1/2}} \exp \left\{- \frac{1}{2} \left(x-\mu_L\right)^{'} \Sigma_L^{-1} \left(x-\mu_L \right)  \right\}\\
M_M(x) &= \mathcal{N}(x|\mu_M,\Sigma_M)\\
&=\frac{1}{2\pi |\Sigma_M|^{1/2}} \exp \left\{- \frac{1}{2} \left(x-\mu_M\right)^{'} \Sigma_M^{-1} \left(x-\mu_M \right)  \right\}\\
M_S(x) &= \mathcal{N}(x|\mu_S,\Sigma_S)\\
&=\frac{1}{2\pi |\Sigma_S|^{1/2}} \exp \left\{- \frac{1}{2} \left(x-\mu_S\right)^{'} \Sigma_S^{-1} \left(x-\mu_S \right)  \right\}\\
\end{aligned}
\end{equation}
where $x\triangleq(BoxS, DtoC)$ the same as before; $S$, $M$ and $L$ in the subscripts denote `small', `middle', and `large', respectively; $\mu$ and $\Sigma$ denote the mean and covariance matrix of the data distribution, which are calculated by maximum likelihood estimation \cite{myung2003tutorial}.


The fuzzy-rule based CKIM is designed as follows:\\
If $x$ matches $M_L$, then object's label is `large' with degree $M_L(x)$;\\
If $x$ matches $M_M$, then object's label is `middle' with degree $M_M(x)$;\\
If $x$ matches $M_S$, then object's label is `small' with degree $M_S(x)$.

A crisp output is then calculated by a defuzzification approach to integrate membership degrees between the object and all rules.
In this paper, the fine-grained result is determined by the Center of Maximum method, defined as follows:

\begin{equation}
Prediction = \frac{\sum_{i}M_{i}(x)y_{i}}{\sum_{i} M_{i}(x)}, i\in\{L,M,S\}
\end{equation}

Note that since these two types of CKIM have few parameters to be optimized, the amount of data required to train them is almost negligible compared to those required to train a DNN model.

\section{Experiments}\label{sec:experiment}
We experimentally evaluated the performance of our proposed CKIM-assisted DNN detector. We compared our method against the SOTA methods, including YOLOv7-tiny and MobileNetv3-SSD, both being developed for resource-constrained scenarios. We integrated CKIM with YOLOv7-tiny and MobileNetv3-SSD and assessed whether it resulted in improved performance.
\subsection{Experimental Setup}
We conducted experiments on the open-source CLEVR data set \cite{johnson2017clevr}. It consists of images containing objects and questions related to them. The objects' attributes include size (big, small), color (brown, blue, cyan, gray, green, purple, red, yellow), material (metal, rubber), and shape (cube, cylinder, sphere). According to these attributes, objects are divided into 96 fine-grained categories. We term this dataset as CLEVR-96.
In our experiment, when using our approach, we removed the size attribute to train a lightweight object detector that considers $96/2=48$ coarse-grained labels at first. Then, for each test image, our approach employs this lightweight object detector to yield category label for it, then invoke the CKIM to infer its size attribute, namely the fine-grained label. To simulate complex environments in the real world, we constructed a new dataset by introducing a collection of objects of middle size. It has 144 fine-grained classes. We term this dataset as CLEVR-144 in what follows. See Fig.\ref{fig:CLEVR} for example pictures in these datasets.
Both CLEVR-96 and CLEVR-144 contain 16,000 images as the training set, 2,000 images as the validation set, and the other 2,000 images as the test set.
\subsection{Performance Metrics}
In our evaluation, we considered three different perspectives: detection accuracy, model size, and processing latency:
\begin{itemize}
  \item Detection Accuracy is measured by the mean Average Precision while IoU=0.5 (mAP@0.5), which is a commonly used metric for evaluating object detectors. It calculates the mean of the average precision over all classifications for every bounding box with an IoU greater than 0.5. Larger mAP@0.5 means higher accuracy.
  \item Model Size is measured by the memory space that the model consumes. This is important in resource-constrained scenarios where memory is limited.
  \item Latency is defined as the average time a method takes to process one image. In our experiments, the time unit was set as millisecond (ms). This is also an important factor to consider in resource-constrained scenarios where real-time processing is required.
\end{itemize}
\begin{table}[h]
\small
\begin{center}
\begin{tabular}{cccc}
\toprule
 &Accuracy $\uparrow$ & Model Size $\downarrow$ & Latency $\downarrow$\\
\hline
MobileNetv3\_SSD & 0.968& 87.61MB& 82\\
Our approach &\textbf{0.978} &\textbf{49.63MB+2KB} & \textbf{60}\\
\hline
YOLOv7-tiny &0.972 &23.31MB& 70 \\
Our approach &\textbf{0.983} &\textbf{22.89MB+2KB} & \textbf{62}\\
\bottomrule
\end{tabular}
\end{center}
\caption{Experiments on the CLEVR-96 dataset. The best performances for each model type are highlighted in \textbf{bold}. In contrast to the fine-grained object detector (MobileNetv3 or YOLOv7-tiny) trained with all 96 fine-grained labels, our approach utilizes a much lighter version of MobileNetv3 or YOLOv7-tiny. This lightweight model is trained with only 48 coarse-grained labels by excluding the size attribute from the dataset. The size-related fine-grained labels of the test images are inferred using the CKIM component of our approach.}
\label{tab:full1}
\end{table}

\begin{table}[h]
\small
\begin{center}
\begin{tabular}{cccc}
\toprule
 &Accuracy $\uparrow$ & Model Size $\downarrow$  & Latency $\downarrow$\\
\hline

MobileNetv3\_SSD & 0.970& 125.59MB& 84\\

Our approach(crisp) & 0.968 &\textbf{49.63MB+2KB} & \textbf{61}\\

Our approach(fuzzy) &\textbf{0.978} &\textbf{49.63MB+2KB} & 62 \\
\hline

YOLOv7-tiny &0.965& 23.73MB & 75\\

Our approach(crisp) & 0.971 &\textbf{22.89MB+2KB} & 64\\

Our approach(fuzzy) &\textbf{0.980} &\textbf{22.89MB+2KB} & \textbf{62}\\

\bottomrule
\end{tabular}
\end{center}
\caption{Experiments on the CLEVR-144 dataset}
\label{tab:full2}
\end{table}
\subsection{Experimental Results}
We trained all the involved models using all 16,000 images in the training set. The performance of these models on the CLEVR-96 and CLEVR-144 datasets are shown in Tables \ref{tab:full1} and \ref{tab:full2}, respectively. We observe that our approach outperforms both baseline methods in terms of Detection Accuracy, Model Size and Latency. Since this dataset only has two fine-grained labels ('small' and 'large'), there is no label ambiguity issue involved. So only the crisp-rule based version of our approach is considered in the experiment. On the CLEVR-144 dataset, which involves three fine-grained labels (`small', `middle', and `large'), Table \ref{tab:full2} shows that the fuzzy-rule based version of our approach is preferable to the crisp-rule based one. This demonstrates the ability of the fuzzy method to deal with semantic ambiguity in the fine-grained labels.

We also evaluated a baseline fine-grained YOLOv4 model. On the CLEVR-96 dataset, this model achieves a high detection accuracy of 0.998, but has a much larger model size of 246.35MB and latency of 176ms than the methods listed in Table \ref{tab:full1}. On the CLEVR-144 dataset, it achieves detection accuracy of 0.998 again, while it also has a much larger model size of 247.34MB and latency of 183ms than the methods listed in Table \ref{tab:full2}.

\subsection{On experiments with 5,000 randomly selected images being used for model training}
In order to simulate cases where the edge device does not have enough memory space to store all training data, we conducted experiments using 5,000 randomly selected images from the training set for model training. The experimental results on the CLEVR-96 and CLEVR-144 datasets are presented in Tables \ref{tab:50001} and \ref{tab:50002}, respectively.

As shown in the tables, for the case with less training data, the benefits given by our proposed method become more remarkable. Our method outperforms all baseline methods according to all performance metrics. In addition, we once again observed the benefits of fuzzy inference on the CLEVR-144 dataset. The fuzzy-based method significantly outperforms the crisp-rule based method in terms of detection accuracy.
\begin{table}[ht]
\small
\centering
\begin{tabular}{cccc}
\toprule

 & Accuracy $\uparrow$ & Model Size $\downarrow$  & Latency $\downarrow$ \\
\hline

MobileNetv3\_SSD &0.902& 87.61MB & 82\\

Our Approach &\textbf{0.912}& \textbf{49.63MB+2KB}& \textbf{60}\\

\hline
YOLOv7-tiny &0.928& 23.31MB& 71 \\
Our Approach &\textbf{0.984}& \textbf{22.89MB+2KB} & \textbf{61} \\

\bottomrule
\end{tabular}
\caption{Fine-grained object detection performance on the CLEVR-96 dataset using 5,000 randomly selected images for model training. All terms are defined in the same way as in Table \ref{tab:full1}. As part of our experiment, we also evaluated a baseline fine-grained YOLOv4 model. This model achieves a high detection accuracy of 0.972, but has a much larger model size of 246.35MB and latency of 178ms than the methods listed here.}
\label{tab:50001}
\end{table}
\begin{table}[ht]
\small
\centering
\begin{tabular}{cccc}
\toprule

 & Accuracy $\uparrow$  & Model Size $\downarrow$  & Latency $\downarrow$\\
\hline
MobileNetv3\_SSD & 0.857& 125.59MB & 83\\

Our Approach (crisp) & 0.862& \textbf{49.63MB}+2KB & 60\\

Our Approach (fuzzy)  &\textbf{0.877} &\textbf{49.63MB}+2KB & \textbf{59}\\

\hline
YOLOv7-tiny &0.865& 23.73MB& 72\\

Our Approach (crisp) &0.892 &\textbf{22.89MB+2KB} & 65\\

Our Approach (fuzzy) & \textbf{0.914} &\textbf{22.89MB+2KB}& \textbf{62}\\

\bottomrule
\end{tabular}
\caption{Fine-grained object detection performance on the CLEVR-144 dataset using 5,000 randomly selected images for model training. As part of our experiment, we also evaluated a baseline fine-grained YOLOv4 model. This model achieves a high detection accuracy of 0.957, but has a much larger model size of 247.34MB and latency of 175ms than the methods listed here.}
\label{tab:50002}
\end{table}

\subsection{Training process comparison for fine-grained and coarse-grained object detectors}
\begin{figure}[ht]
\centering
\includegraphics[width=0.4\textwidth]{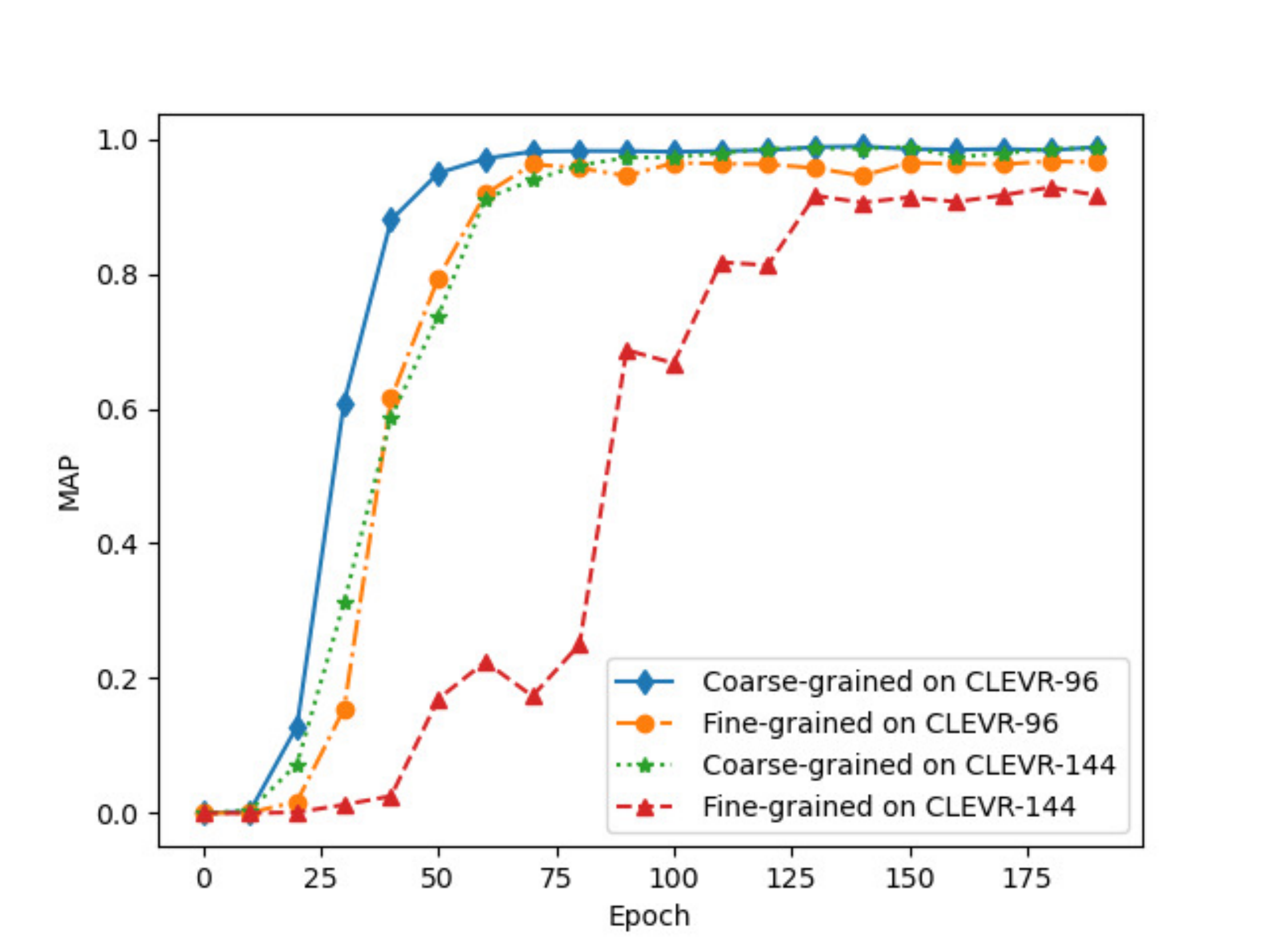}
\caption{Training process of a pair of fine-grained and coarse-grained object detectors}
\label{fig:train_5000}
\end{figure}

In order to investigate the difference in the training process between fine-grained and coarse-grained object detectors, we conducted an experiment using the YOLOv7-tiny model. Specifically, we compared the training process of a YOLOv7-tiny based fine-grained detector and a YOLOv7-tiny based coarse-grained detector. The results, shown in Fig.\ref{fig:train_5000}, demonstrate that the coarse-grained detector converges much faster than its fine-grained counterpart on both the CLEVR-96 and CLEVR-144 datasets.

Since our proposed fine-grained object detector consists of a coarse-grained detector and a lightweight CKIM, the convergence speed of our method is almost the same as the coarse-grained detector it employs. This indicates that the convergence speed of our proposed fine-grained detector is much faster than its counterpart fine-grained detector.

Overall, these results demonstrate that our proposed method achieves efficient and effective fine-grained object detection by leveraging a lightweight commonsense knowledge inference module with a coarse-grained object detector, achieving high accuracy while maintaining fast convergence times.



\section{Conclusions}\label{sec:conclusions}
In this paper, we focused on a particular fine-grained object detection problem: efficiently identifying the category of each object and determining its size in images containing multiple objects of the same category but varying in sizes. Unlike typical fine-grained object detectors that rely on fine-grained data annotations, our approach does not require such detailed annotations. Furthermore, we demonstrated that our approach is lightweight, faster, and more computationally efficient compared to conventional fine-grained detectors, while maintaining higher detection accuracy.

\begin{figure}[th]
\centering
\includegraphics[width=0.45\textwidth]{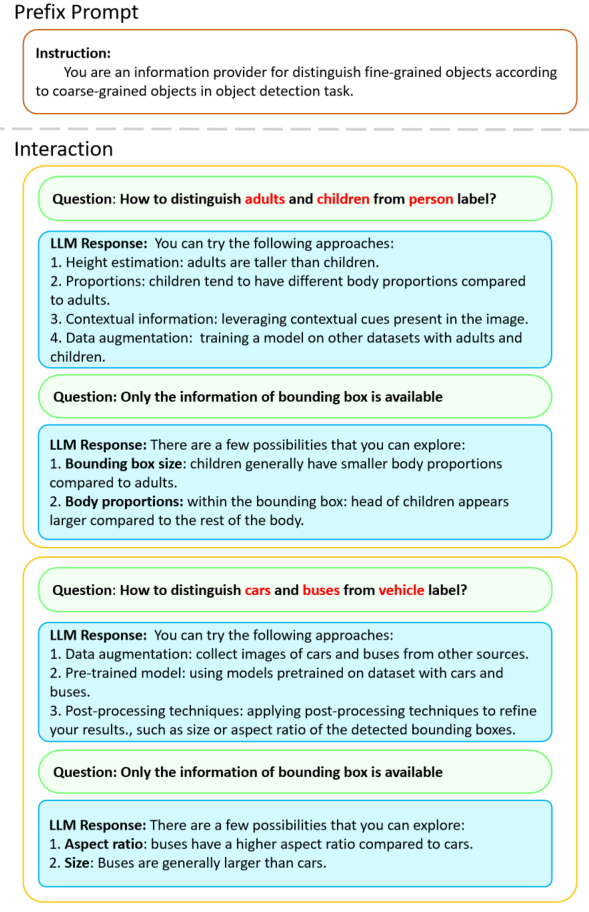}
\caption{Example shows of prompt designs for an LLM that automatically provides knowledge to help distinguish between fine-grained and coarse-grained labels. The prefix prompt offers task instructions, and during the interaction with LLM, only fine-grained and coarse-grained labels need to be provided. Moreover, the attributes used for distinguishing fine-grained labels can be generated automatically.}
\label{fig:LLM_prompt}
\end{figure}

One may question how to acquire commonsense knowledge. Frequently, it is derived from the experience or intuition of human experts. Alternatively, it can also be obtained from a large language model, as preliminarily. To explore the potential of using LLMs to generate commonsense knowledge, we conducted several prompt designs utilizing the Vicuna-7B model, a commonly used open-source LLM model. The response of the LLM can be observed in Fig. \ref{fig:LLM_prompt}. The results demonstrate that LLMs are indeed capable of generating significant commonsense knowledge.

\bibliographystyle{IEEEtran}
\bibliography{ref}

\begin{thebibliography}{10}
\providecommand{\url}[1]{#1}
\csname url@samestyle\endcsname
\providecommand{\newblock}{\relax}
\providecommand{\bibinfo}[2]{#2}
\providecommand{\BIBentrySTDinterwordspacing}{\spaceskip=0pt\relax}
\providecommand{\BIBentryALTinterwordstretchfactor}{4}
\providecommand{\BIBentryALTinterwordspacing}{\spaceskip=\fontdimen2\font plus
\BIBentryALTinterwordstretchfactor\fontdimen3\font minus
  \fontdimen4\font\relax}
\providecommand{\BIBforeignlanguage}[2]{{%
\expandafter\ifx\csname l@#1\endcsname\relax
\typeout{** WARNING: IEEEtran.bst: No hyphenation pattern has been}%
\typeout{** loaded for the language `#1'. Using the pattern for}%
\typeout{** the default language instead.}%
\else
\language=\csname l@#1\endcsname
\fi
#2}}
\providecommand{\BIBdecl}{\relax}
\BIBdecl

\bibitem{redmon2018yolov3}
J.~Redmon and A.~Farhadi, ``Yolov3: An incremental improvement,'' \emph{arXiv
  preprint arXiv:1804.02767}, 2015.

\bibitem{ren2015faster}
S.~Ren, K.~He, R.~Girshick, and J.~Sun, ``Faster {R-CNN}: Towards real-time
  object detection with region proposal networks,'' \emph{Advances in Neural
  Information Processing Systems}, vol.~28, 2015.

\bibitem{zaidi2022survey}
S.~Zaidi, M.~Ansari, A.~Aslam, N.~Kanwal, M.~Asghar, and B.~Lee, ``A survey of
  modern deep learning based object detection models,'' \emph{Digital Signal
  Processing}, p. 103514, 2022.

\bibitem{chen2021decision}
T.~Chen, C.~Shang, P.~Su, E.~Keravnou-Papailiou, Y.~Zhao, G.~Antoniou, and
  Q.~Shen, ``A decision tree-initialised neuro-fuzzy approach for clinical
  decision support,'' \emph{Artificial Intelligence in Medicine}, vol. 111, p.
  101986, 2021.

\bibitem{carter2021fuzzy}
J.~Carter, F.~Chiclana, A.~S. Khuman, and T.~Chen, ``Fuzzy logic: Recent
  applications and developments,'' 2021.

\bibitem{redmon2016you}
J.~Redmon, S.~Divvala, R.~Girshick, and A.~Farhadi, ``You only look once:
  Unified, real-time object detection,'' in \emph{Proc. of the IEEE/CVF Conf.
  on Computer Vision and Pattern Recognition}, 2016, pp. 779--788.

\bibitem{wang2021scaled}
C.~Wang, A.~Bochkovskiy, and H.~M. Liao, ``Scaled-yolov4: Scaling cross stage
  partial network,'' in \emph{Proc. of the IEEE/CVF Conf. on Computer Vision
  and Pattern Recognition}, 2021, pp. 13\,029--13\,038.

\bibitem{wang2022yolov7}
------, ``Yolov7: Trainable bag-of-freebies sets new state-of-the-art for
  real-time object detectors,'' \emph{arXiv preprint arXiv:2207.02696}, 2022.

\bibitem{liu2016ssd}
W.~Liu, D.~Anguelov, D.~Erhan, C.~Szegedy, S.~Reed, C.~Fu, and A.~C. Berg,
  ``S{SD}: Single shot multibox detector,'' in \emph{Proc. of European Conf. on
  Computer Vision}, 2016, pp. 21--37.

\bibitem{howard2017mobilenets}
A.~G. Howard, M.~Zhu, B.~Chen, D.~Kalenichenko, W.~Wang, T.~Weyand,
  M.~Andreetto, and H.~Adam, ``Mobilenets: Efficient convolutional neural
  networks for mobile vision applications,'' \emph{arXiv preprint
  arXiv:1704.04861}, 2017.

\bibitem{sandler2018mobilenetv2}
M.~Sandler, A.~Howard, M.~Zhu, A.~Zhmoginov, and L.~Chen, ``Mobilenetv2:
  Inverted residuals and linear bottlenecks,'' in \emph{Proc. of the IEEE/CVF
  Conf. on Computer Vision and Pattern Recognition}, 2018, pp. 4510--4520.

\bibitem{howard2019searching}
A.~Howard, M.~Sandler, G.~Chu, L.~Chen, B.~Chen, M.~Tan, W.~Wang, Y.~Zhu,
  R.~Pang, and V.~Vasudevan, ``Searching for mobilenetv3,'' in \emph{Proc. of
  the IEEE/CVF Inter. Conf. on Computer Vision}, 2019, pp. 1314--1324.

\bibitem{KURFESS2003609}
\BIBentryALTinterwordspacing
F.~J. Kurfess, ``Artificial intelligence,'' in \emph{Encyclopedia of Physical
  Science and Technology (Third Edition)}, third edition~ed., R.~A. Meyers,
  Ed.\hskip 1em plus 0.5em minus 0.4em\relax New York: Academic Press, 2003,
  pp. 609--629. [Online]. Available:
  \url{https://www.sciencedirect.com/science/article/pii/B0122274105000272}
\BIBentrySTDinterwordspacing

\bibitem{zadeh1973outline}
L.~A. Zadeh, ``Outline of a new approach to the analysis of complex systems and
  decision processes,'' \emph{IEEE Transactions on systems, Man, and
  Cybernetics}, no.~1, pp. 28--44, 1973.

\bibitem{Scherer2012}
R.~Scherer, \emph{Multiple fuzzy classification systems}.\hskip 1em plus 0.5em
  minus 0.4em\relax Springer, 2012, vol. 288.

\bibitem{yang2022neurosymbolic}
K.~Yang, ``Neurosymbolic machine learning for reasoning,'' Ph.D. dissertation,
  Princeton University, 2022.

\bibitem{garcez2022neural}
A.~Garcez, S.~Bader, H.~Bowman, L.~C. Lamb, L.~de~Penning, B.~Illuminoo,
  H.~Poon, and C.~G. Zaverucha, ``Neural-symbolic learning and reasoning: A
  survey and interpretation,'' \emph{Neuro-Symbolic Artificial Intelligence:
  The State of the Art}, vol. 342, no.~1, p. 327, 2022.

\bibitem{yu2023survey}
D.~Yu, B.~Yang, D.~Liu, H.~Wang, and S.~Pan, ``A survey on neural-symbolic
  learning systems,'' \emph{Neural Networks}, 2023.

\bibitem{mikolov2013distributed}
T.~Mikolov, I.~Sutskever, K.~Chen, G.~S. Corrado, and J.~Dean, ``Distributed
  representations of words and phrases and their compositionality,''
  \emph{Advances in Neural Information Processing Systems}, vol.~26, 2013.

\bibitem{pennington2014glove}
J.~Pennington, R.~Socher, and C.~D. Manning, ``Glove: Global vectors for word
  representation,'' in \emph{Proc. of Conf. on Empirical Methods in Natural
  Language Processing (EMNLP)}, 2014, pp. 1532--1543.

\bibitem{nickel2016holographic}
M.~Nickel, L.~Rosasco, and T.~Poggio, ``Holographic embeddings of knowledge
  graphs,'' in \emph{Proc. of the AAAI Conf. on artificial intelligence},
  vol.~30, no.~1, 2016.

\bibitem{yi2018neural}
K.~Yi, J.~Wu, C.~Gan, A.~Torralba, P.~Kohli, and J.~Tenenbaum,
  ``Neural-symbolic vqa: Disentangling reasoning from vision and language
  understanding,'' \emph{Advances in neural information processing systems},
  vol.~31, 2018.

\bibitem{eiter2022neuro}
T.~Eiter, N.~Higuera, J.~Oetsch, and M.~Pritz, ``A neuro-symbolic asp pipeline
  for visual question answering,'' \emph{Theory and Practice of Logic
  Programming}, vol.~22, no.~5, pp. 739--754, 2022.

\bibitem{lavalley2008logistic}
M.~P. LaValley, ``Logistic regression,'' \emph{Circulation}, vol. 117, no.~18,
  pp. 2395--2399, 2008.

\bibitem{2021Approximate}
F.~Li, C.~Shang, Y.~Li, J.~Yang, and Q.~Shen, ``Approximate reasoning with
  fuzzy rule interpolation: background and recent advances,'' \emph{Artificial
  Intelligence Review}, vol.~54, no.~6, pp. 4543--4590, 2021.

\bibitem{kreinovich1992gaussian}
V.~Kreinovich, C.~Quintana, and L.~Reznik, ``Gaussian membership functions are
  most adequate in representing uncertainty in measurements,'' in
  \emph{Proceedings of NAFIPS}, vol.~92, 1992, pp. 15--17.

\bibitem{myung2003tutorial}
I.~J. Myung, ``Tutorial on maximum likelihood estimation,'' \emph{Journal of
  mathematical Psychology}, vol.~47, no.~1, pp. 90--100, 2003.

\bibitem{johnson2017clevr}
J.~Johnson, B.~Hariharan, L.~Van Der~Maaten, L.~Fei-Fei, C.~Lawrence~Zitnick,
  and R.~Girshick, ``C{LEVR}: A diagnostic dataset for compositional language
  and elementary visual reasoning,'' in \emph{Proc. of the IEEE Conf. on
  Computer Vision and Pattern Recognition}, 2017, pp. 2901--2910.

\end{thebibliography}
\end{document}